# A comparative study of statistical and machine learning models on near-real-time daily $CO_2$ emissions prediction.

Xiangqian Li

School of Statistics, Capital University of Economics and Business, Beijing, P.R. China

## Abstract

The rapid ascent in carbon dioxide emissions is a major cause of global warming and climate change, which pose a huge threat to human survival and impose far-reaching influence on the global ecosystem. Therefore, it is very necessary to effectively control carbon dioxide emissions by accurately predicting and analyzing the change trend timely, so as to provide a reference for carbon dioxide emissions mitigation measures. This paper is aiming to select a suitable model to predict the near-real-time daily $CO_2$ emissions based on univariate daily time-series data from January 1st, 2020 to September 30st, 2022 of all sectors (Power, Industry, Ground Transport, Residential, Domestic Aviation, International Aviation) in China. We proposed six prediction models, which including three statistical models: Grey prediction (GM(1,1)), autoregressive integrated moving average (ARIMA) and seasonal autoregressive integrated moving average with exogenous factors (SARIMAX); three machine learning models: artificial neural network (ANN), random forest (RF) and long short term memory (LSTM). To evaluate the performance of these models, five criteria: Mean Squared Error (MSE), Root Mean Squared Error (RMSE), Mean Absolute Error (MAE), Mean Absolute Percentage Error (MAPE) and Coefficient of Determination ($R^2$) are imported and discussed in detail. In the results, three machine learning models perform better than that three statistical models, in which LSTM model performs the best on five criteria values for daily $CO_2$ emissions prediction with the 3.5179e-04 MSE value, 0.0187 RMSE value, 0.0140 MAE value, 14.8291% MAPE value and 0.9844 $R^2$ value. Therefore, LSTM model is suggested as one of the most appropriate model for near-real-time daily $CO_2$ emissions prediction based on long daily time series data. In the end, according to the results of this study, a few policy recommendations are presented to reduce carbon emissions by various departments in China.

# 1. Introduction

Carbon dioxide ($CO_2$), which makes up 1% of the atmospheric gases but occupies about 81% of the total emissions gases, is a main component of Green House Gas (GHG) emissions [5]. Carbon dioxide emissions can lead to climate change and threat the various aspects of human life ([22]; [17]). The problem of global climate change, particularly global warming, is becoming more and more serious, among which carbon dioxide is the main cause of climate change ([6]; [44]; [26]). As the problem of global climate deterioration increasing serious, countries all over the world are actively participating in dealing with climate challenge. As the world's largest $CO_2$ emitter, China's carbon dioxide emissions exceeded 11.9 billion tons in 2021, accounting for 33 percent of the global total [25], hence China is facing great pressure to reduce carbon emissions, and mitigating carbon emissions is becoming more and more imminent. China has implemented $CO_2$ emissions control through a series of international commitments and national emission targets in recent years. At the 75th United Nations General Assembly in September 2020, President Xi Jinping announced that China will achieve carbon peak by 2030 and carbon neutral by 2060.

The premise of carbon emissions mitigation targets and policies is the high precision prediction results as a reference, so the research on $CO_2$ prediction becomes particularly important and has became a worldwide concern [1]. As far as we know, most of the researches on prediction of $CO_2$ emissions have focused on annual emissions prediction. Although the annual emissions prediction could provide references for the formulation and implementation of long-term carbon mitigation policies. but because of the acquisition of annual emissions data often has a certain lag and limited sample size as well as the annual emissions prediction cannot capture small changes in the short-term based on annual data. Therefore, the ability of carbon emissions prediction based on annual emissions data is not accurate enough. So, in order to monitor the dynamic trend of $CO_2$ emissions and set short-term emission mitigation targets timely, near-real-time daily predictions play a vital role, which could shorten the response time of policy adjustments. In order to make up the gap of existing research on short-term daily $CO_2$ emissions prediction, we propose three statistical models (GM(1,1), ARIMA, and SARIMAX), three machine learning models (ANN, RF, and LSTM) for daily $CO_2$ emissions prediction and also carefully analysis the performance criteria based on five criteria (MSE, RMSE, MAE, MAPE, and $R^2$) to get the best performing model. The dataset used in this paper is from January 1st, 2020 to September 30st, 2022 of total 1004 daily carbon emissions data in China.

The main innovations and contributions of this research are as follows: (1) An extended near-real-time $CO_2$ emissions dataset with daily frequency are used to selected the best suited model for daily prediction, which can improve the accuracy of prediction. (2) This paper fills the gap in daily prediction of carbon dioxide emissions which will aid in decision-making by policymakers and governments for timely policy adjustment. (3) This paper has made a comparative analysis of different models and their performance in predicting $CO_2$ emissions, which would be helpful for other researchers to select the best suitable prediction models for daily prediction.

The remaining structure of this article is organized as follows: Section 2 is "Literature review", which summarize the work in the field of carbon dioxide emissions prediction and related works. Section 3 is "Methodology", which introduces data sources, proposed models and corresponding evaluation criteria used to evaluate prediction models. Section 4 is "Results and discussions", in which the performance of each model is evaluated and compared. Section 5 is "Conclusions and policy implications", which includes conclusions, policy recommendations, flaws, and future research plans.

## 2. Literature review

Many scholars devoted to the $CO_2$ emissions prediction based on time series data, which can be divided into two categories: influencing factor prediction and time series prediction. In the first category, influencing factors that have a strong correlation with carbon emissions are chosen as the inputs of the model to predict the future carbon emissions. In the second category, univariate time-series history data are used as inputs of the model to predict the future trends in carbon dioxide emissions.

There have been a lot of papers to predict carbon dioxide emissions suing a variety of prediction models. In the beginning, the carbon emissions prediction methods are mainly based on traditional statistical prediction models, such as GM, ARIMA, SARIMAX and so on. As the time series data are highly nonlinear, non-stationary and enlarged, traditional statistical models are unable to deal with this type problem effectively [60]. Therefore, scholars have begun to draw into the machine learning (containing deep learning) prediction models, such as ANN, RF, SVM, LSTM etc, to deal with nonlinear time series data. Some representative models for predicting are briefly reviewed in the following.

Grey prediction model (GM) which was first proposed by [14] is a method for predicting systems with insufficient information. In [52], the authors proposed a nonlinear Grey model to predict the future carbon emissions from fossil energy consumption in China from 2014 to 2020 and the results showed that higher prediction accuracy compared with the traditional GM and ARIMA model. [21] proposed a GM model to predict the energy related $CO_2$ emissions in

Turkey based on the annual emissions data from 1965-2012. Meanwhile, the predicting performance of the model was examined by three indexes (MAE, MAPE, and the post error ratio (C) ) and results indicated that the GM model can be used as a potential prediction model for $CO_2$ emissions. [16] advanced an improved GM model (denoted by TDVGM(1,4)) and compared with four models (GM(1,4), DGM(1,4), ARIMA and MLRM) by using three evaluation indexes (MAPE, RMSE and FD). In summary, comparing results indicated that TDVGM(1,4) had the best reliable performance in $CO_2$ emissions prediction. To improve GM model's performance, [37] designed a integrated predicting model with GM(1,1), ARIMA and SOPR, which improved the prediction accuracy by Particle Swarm Optimization. In similar studies, GM model has been applied in different branches such as pollutant emissions, electricity demand, energy consumption etc. Such as in [43], GM model was applied to search the relationship between pollutant emissions, energy consumption, and GDP and then predicted them during 2008-2013 for Brazil based on the annual data from 1980-2007. In the same time, the predictive ability of GM was compared with ARIMA model and all of them had a strong predicting performance with MAPEs of less than 3%. The work by [3] proposed a GM with rolling mechanism (GPRM) model to predict total and industrial power consumption in Turkey, in which this model performed more accurate results than MAED. In similar, there are a large number of studies ([55]; [54]; [59]; [18]; [51]) used GM model for power demand predicting. However, [23] pointed out that the prediction accuracy of GM for time series with volatility is low.

Autoregressive integrated moving average (ARIMA) model has also been used to predict $CO_2$ emissions based on historical data. By [42], they used the Box-Jenkins ARIMA(2, 2, 0) model to predict $CO_2$ emissions of India based on the annual time series data from 1960 to 2017. By relevant experiments, the results implied that the proposed model was reliable and stable. In [30], ARIMA model was proposed to predict $CO_2$ emissions for the period of 2015-2027, which based on the annual times series data from 1980 to 2016 of South Africa. With the lowest RMSE value 0.052072, it is indicated that the ARIMA(4, 2, 3) is the best suitable predicting model in this study. [4] proposed a comparative study of $CO_2$ emissions prediction using ARIMA, HWES and ANN models based on annual time series data in the Gulf countries. In light of the accuracy measures with MAE, MSE, RMSE and MAPE, the study's results showed that the two statistical models ARIMA and HWES do not outperform the ANN model in predicting. ARIMA model has also been used to predict the electricity prices, air quality or pollutants using time series analysis. For example, [13] compared the predicting abilities of univariate models AR and ARIMA with MAE and RMSE performance metrics based on hourly electricity spot prices data from the Leipzig power exchange. The performance of univariate prediction methods such as ARIAM, VAR models had been evaluated based on a number of metrics and it had been observed that the ARIMA model performed better than other models ([20]; [45]).

SARIMA model is an ARIMA model capable of dealing with the seasonality of the dataset, and SARIMAX is SARIMA with X factor, which is nothing but exogenous factors that affect the data [31]. [40] analysed and predicted time series using the SARIMA and SARIMAX models with decomposition method and compared them on performance metric MAPE value where SARIMAX with decomposition model performed the best. The paper of [2] was aimed to compare four predicting models ARIMA, SARIMA, SARIMAX, and multiple linear regression with WAE and RMSE values as the performance metrics to select the best model for predicting the electricity prices of Germany. It was observed that the SARIMAX model performed the best among other models in this paper. Classic ARIMA, SARIMA and SARIMAX models were also used to predict the tree sap flux in sapwood in [58] and the performance of these models in the article was assessed with six statistical criteria, which showed that SARIMAX model performed best among other models on prediction accuracy.

Generally speaking, GM, ARIMA and SARIMAX models, as classical statistical models, are often used in time series prediction. However, these methods have two weakness: (1) They are suitable for stationary time series data, but not suitable for nonlinear cases. (2) It is suitable for predicting based on small sample size, and the accuracy of prediction large sample size is not high. In addition, with the development of computer science, the machine learning (including deep learning) methods for predicting have been favored by researchers due to its strong nonlinear fitting ability, robustness and self-learning ability.

Artificial neural network (ANN) is a method to solve complex problems. ANN can learn from samples and deal with noisy or nonlinear problems, and once trained, they can make predictions efficiently [28]. In the past years, the literature regarding carbon dioxide emissions prediction based on ANN models quickly extended with the development of computer technology. To predict the OPEC $CO_2$ emissions from petroleum consumption, [10] established an HCSANN model based on CS, ANN, and APSO which predicted OPEC $CO_2$ emissions for 3, 6, 9, 12 and 16 years with an improved accuracy and speed over other meta-heuristic models. [9] investigated the $CO_2$ emissions, energy consumption and economic growth prediction based on GMM and ANN models over the period 1985-2013 in Eastern Asian countries. The performance evaluation results showed that ANN model is a suitable model for future $CO_2$ emissions prediction and reveals several interesting and useful implications. In [8], four machine learning models (ANN, SMRGT, SVM and ANFIS) and one statistical model MLR were developed for $CO_2$ emissions prediction of transportation sector. As a result, the best performance model is ANN compared with other models. The study of [39] for India used a machine learning model ANN to predict the $CO_2$ emissions and growth by taking energy consumption as the input, and tried to understand the role of fossil fuel in the determination of $CO_2$ emissions. An novel integrated multi-layer ANN and Bees Algorithm (denoted by BA model) was proposed by [7], which is presented for predicting world's $CO_2$ emissions and energy demand based on twenty seven years data from 1980 to 2006. The main subject in [46] is to predict the GHGs of Turkey using sectoral energy consumption data by ANN model. The results showed that the ANN model performed excellent and it can be used to determine the future level of GHG in Turkey and to take measures to control total emissions. ANN model also had been used to solve the prediction problem on other fields and performed outstanding ([29]; [38]; [47]).

Random forest (RF) is a ensemble method which grow trees as the base learners and combine their predictions by averaging. In particular, RF performs well in high-dimensional settings. In the study of [48], A hybrid model RF-MFO-ELM combining RF, MFO and ELM was proposed for the prediction of carbon dioxide emissions, in which RF was used for influential factors analysis and ELM was used for the prediction. Taking the data of Hebei Province in China from 1995 to 2015 as an example, the results showed that the proposed model outperformed other parallel models in $CO_2$ prediction. In [41], the amounts of $CO_2$, $CH_4$, $N_2O$ changes in Turkey had been predicted according to vehicle types of vehicle mobility on highways using machine learning Linear Regression, Bayesian Ridge, RF Regression, MLP Regression and SVR models. The results concluded that RF model performed the best on the $CO_2$ emissions prediction with the evaluated metrics MAE, MSE, RMSE and $R^2$. [19] aimed at developing an optimal prediction model to predict the accurate of an infectious diarrhea epidemic. An RF model was proposed and compared with classical statistical models ARIMA and SARIMAX, in which the RF model outperformed the ARIMA model with a MAPE of approximately 20% and the performance of the SARIMAX model was compared to that of the ARIMA model with a MAPE reaching approximately 30%.

LSTM, as a kind of RNN, model can effectively use history information to predict the future time point. So it has been widely applied in various scenarios with its excellent performance in long time series prediction. Such as [12] proposed a novel architecture for emergency event prediction based on LSTM model. In the same time, A comparative analysis was proposed to show the effectiveness of the LSTM model compared with three traditional statistical prediction models (MA, WMA, and ARIMA) and three machine learning models (DT, SVM, and MLP) through the evaluation metrics based on historical emergency event data in Guatemala. For the emergency event prediction problem, the LSTM model was superior to other models. [61] innovatively proposed an integrated model LSTM-STIRPAT, which combining the LSTM and STIRPAT models, to predict the $CO_2$ emissions and to identify its influencing factors in 30 provinces in China based on the panel data from 1991 to 2017. The research in [5] was aimed at predicting the $CO_2$ levels in the year 2020 in India using a hybrid deep learning model CNN-LSTM, consisting of CNN and LSTM. Compared with other exponential smoothing models with performance metrics RMSE and MAPE, the CNN-LSTM model performed the best, so that the proposed model can be used in the future to predict $CO_2$ emissions as well as other pollutant levels such as PM, NOx, CO, $SO_2$ etc. An LSTM model was established by [24] to predict China's carbon emissions based on the annual data from 1990 to 2016. In this paper, through the prediction of China's carbon emissions in 2012-2016, and comparing with the Gaussian process regression (GPR) and back propagation neural network (BPNN) models' prediction results, it was shown that the minimum MAPE value of LSTM model indicated that the LSTM model was better than that of BPNN and GPR in $CO_2$ emissions prediction. To compare the performance of statistical and machine learning models for $CO_2$ emissions prediction, [31] used three statistical models (ARIMA, SARIMAX, and Holt-Winters) and three machine learning models (LR, RF, and LSTM) to select the best predicting model based on univariate time series data from 1980 to 2019 in India. Through nine evaluation metrics, LSTM performed the best among all the models in this paper and was supposed as the most appropriate prediction model for $CO_2$ emissions.

Through reviewing the literature about predicting models of $CO_2$ or other gases emissions, we have observed that prediction models can be divided into statistical and machine learning (including deep learning) models. The benefit of statistical models for prediction is that they will work well when the sample size of data is minimal and the data is stable. Also the number of statistical models' parameters that must be determined is quite few compared with machine learning models. However, such statistical models lack a few key power involved with complex predicting tasks based on long and fluctuate time series data. Furthermore, we have observed almost all the researches are based on annual dataset for predicting the next few years' values which we can see as long-term prediction. To fill the gap on the short-term prediction based on univariate daily $CO_2$ emissions data, we will applied three statistical and three machine learning models for effective prediction of $CO_2$ emissions, and get the best performing model among them under the evaluation of five performance criteria.

## 3. Methodology
### 3.1 Data desccription
The necessary data for this paper is collected from Carbon Monitor project (https://carbonmonitor.org) which contains 1004 daily near-real-time time series of China's carbon dioxide emissions from January 1st, 2020 to September 30st, 2022, as shown in Fig.1. The daily carbon dioxide emissions data are calculated as the sum of six sectors (Power, Industry, Ground Transport, Residential, Domestic Aviation, and International Aviation) of China based on the near-real-time activity data ([35]; [36]). In this study, we will use the univariate daily time series carbon dioxide emissions data for the model training and predicting. As we can see from Fig.1, the data used in this paper is somewhat seasonal and daily $CO_2$ emissions peak in January and trough in February every year. At the same time, it can be seen that due to the impact of the initial outburst of COVID-19, carbon dioxide emissions during the period from January 1st, to February 11st, 2020 decreased significantly compared with 2021 and 2022.

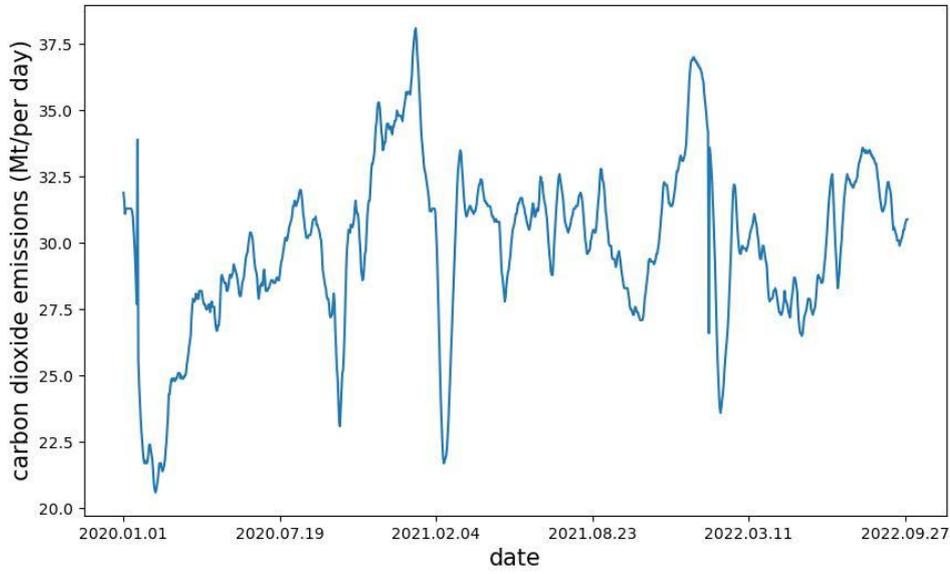

Fig.1. $CO_2$ emissions (Mt/per day) in China from January 1st, 2020 to September 30st, 2022.

The descriptive statistical analysis of the $CO_2$ emissions data is shown in Table 1, which count, maximum, minimum, mean, median, range, skewness, kurtosis, standard deviation and standard error values of the data are calculated. Specially, the difference between the standard error and standard deviation values can be associated with the difference in the size of the numerical values of the data.

Table 1. Descriptive Statistics of Data

| Label | Count | Maximum | Minimum | Mean | Median | Range | Skewness | Kurtosis | Standard Deviation | Standard Error | Total |
|---|---|---|---|---|---|---|---|---|---|---|---|
| $CO_2$ emissions | 1004 | 38.1 | 20.6 | 29.89 | 30.2 | 17.5 | -0.42 | 0.68 | 3.14 | 0.10 | 30013.4 |

## 3.2 Work frame

The research process is divided into three parts and the research framework of this paper is shown in Figure 2.

The first part is data preparation. Firstly, we pre-process the collected univariate time series data, including data normalization, data adjustment to model input format, etc. Then, the data are divided into training sets (80%), validation sets (10%) and test sets (10%), in which the training sets are used to train the model and determine the model parameters; the validation sets are used to determine the structure and adjust the hyperparameters of the model; the test sets are used to verify the generalization ability of the model. Finally, the processed data are passed to the model.

The second part is the model building. This paper considers six models: three statistical models (GM(1,1), ARIMA, and SARIMAX) and three machine learning models (ANN, RF, and LSTM). We will use ModelCheckPoint and GridSearch technologies to seek the best model's parameters when the pre-processed training and verification sets are trained for each model respectively.

The third part is model selection. The six trained models are tested on the test set, and comprehensive evaluation is carried out according to the five evaluation criteria (MSE, RMSE, MAE, MAPE, and $R^2$). Finally the best performing model will be selected and used for $CO_2$ emissions prediction according to comprehensive evaluation.

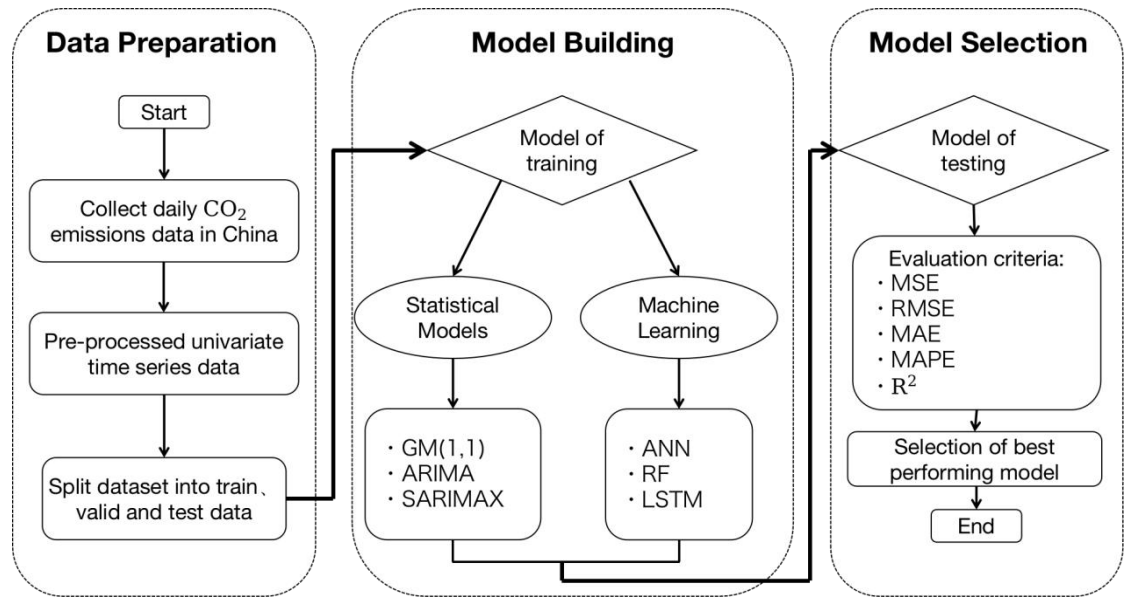

Fig.2. Flowchart of the research in this paper.

## 3.3 Model

### 3.3.1 Grey Model (GM(1,1))

Grey model usually denoted by GM (p,q) where p is the order of the differential equation and q is the number of variables [15]. The most widely used version of Grey model is GM (1,1) model, which will be introduced as follows:

Assuming that the primitive time series data $X^{(0)}$, denoted by

$$X^{(0)} = (x_1^{(0)}, x_2^{(0)}, \ldots, x_n^{(0)}) \qquad (1)$$

where n is the size of data.

The accumulated sequence $X^{(1)}$, defined by the first-order accumulated generating operation of $X^{(0)}$, denoted by

$$X^{(1)} = (x_1^{(1)}, x_2^{(1)}, \ldots, x_n^{(1)}) \qquad (2)$$

where $x_j^{(1)} = \sum_{i=1}^{j} x_i^{(0)}$, $j = 1, 2, 3, \ldots, n$.

The generated adjacent mean sequence $Z^{(1)}$ of $X^{(1)}$ is denoted by

$$Z^{(1)} = (z_1^{(1)}, z_2^{(1)}, \ldots, z_n^{(1)}) \tag{3}$$

where $z_j^{(1)} = \frac{x_j^{(1)} + x_{j+1}^{(1)}}{2}$, $j = 1, 2, 3, \ldots, n$.

Then

$$x_j^{(0)} + az_j^{(1)} = b, j = 1, 2, 3, \ldots, n. \tag{4}$$

is called a GM(1,1) model where -a and b are called the development coefficient and grey action quantity, respectively. Then corresponding whitening equation is:

$$\frac{dx_t^{(1)}}{dt} + ax_t^{(1)} = b \tag{5}$$

By using least-squares method and initial value $x_1^{(1)} = x_1^{(0)}$, the solution of $x_j^{(0)}$ at time j in equation (5) is:

$$x_j^{(1)}(p) = (x_1^{(0)} - \frac{b}{a})e^{-a(j-1)} + \frac{b}{a}, j = 2, 3, \ldots, n. \tag{6}$$

To get the predicted value of the primitive data at time k, the first-order inverse accumulated generating operation is used to obtain the modeling value:

$$x_j^{(0)}(p) = (x_1^{(0)} - \frac{b}{a})e^{-a(j-1)}(1 - e^a), j = 2, 3, \ldots, n. \tag{7}$$

### 3.3.2 Autoregressive Integrated Moving Average（ARIMA）

ARIMA model analyzes and normalizes unsteady time series by adding another process to the ARMA model, where the ARMA model is a mixture of autoregressive model (AR) and moving average model (MA). ARIMA model is one of the most common statistical models used to predict time series data. This model contains three parameters p, d and q, usually denoted by ARIMA (p,d,q) where p represents the lags of the time series data itself used in the prediction model, also known as Auto-Regressive (AR) term; d represents the order of difference that makes the time series data stable, also known as Integrated (I) term; q represents the lags number of prediction errors adopted in the prediction model, also known as Moving Average (MA) term. The mathematical expression of ARIMA model is:

$$\phi_p(B)(1-B)^d y_t = \theta_q(B)\varepsilon_t \tag{8}$$

where $y_t$ is the original time series data; $\phi_p$ is the autoregressive coefficient; B represents the backward shift operator; d denotes the order of difference; $\theta_q$ is the moving average coefficient; $\varepsilon_t$ represents a white noise sequence or error term and it follows a normal distribution with constant variance and zero mean [11]. In particular, ARIMA(p,0,0) is the AR model and ARIMA(0,0,q) is the MA model.

The modeling process of ARIMA is as follows:

Step 1: Check the stationarity of the time series. By observing the time series diagram and unit root test (ADF), we can determine whether the time series is stationary.

Step 2: Stabilize the sequence. For a non-stationary sequence, it can be made stationary by difference operation until the sequence is a non-white noise sequence after d differences.

Step 3: Determine the order of the model. The appropriate autocorrelation order p and moving average order q are selected according to the autocorrelation and partial autocorrelation of samples, and then the model is fitted.

Step 4: Estimate model's parameters. The least square method is used to estimate the regression coefficient of the sequence.

Step 5: Test the model. The residual white noise test and parameter type test are carried out to judge whether the model is reasonable. If the residual sequence is not white noise sequence, then go back to step 3 to re-model until the parameter test and residual white noise test of the model are passed.

Step 6: Prediction by the model. The ARIMA model which passes the parameter test and residual white noise test is used to predict the time series.

The advantage of the ARIMA model is its simplicity and the disadvantage is that the model requires the time series to be stable, or stable after difference. Specially, the model has a good fitting effect on linear data, but it is not sensitive to nonlinear data and the prediction effect is poor. In this paper, we determine the use of ARIMA (0,1,3) for $CO_2$ emissions prediction through GridSearch.

### 3.3.3 Seasonal Autoregressive-Integrated Moving Average With Exogenous Factors (SARIMAX)

If the time series data have seasonal fluctuations, then seasonal differences can be added to the ARIMA model to remove them, denoted as SARIMA. The SARIMA model contains the following parameters, non-seasonal parameters: p is the non-seasonal autoregressive parameter, d is the difference fraction, q is the moving average parameter; seasonal parameters: P is the seasonal regression parameter, D is the seasonal difference, Q is the seasonal moving average parameter, and s is the season length. So SARIMA is usually denoted by SARIMA(p,d,q)(P, D, Q)$_s$ [27]. SARIMAX model adds additional exogenous variables on the basis of SARIMA model, and its mathematical expression is as follows:

$$\Phi_p(B^s)\phi_p(B)(1-B^s)^D(1-B)^d y_t = \Theta_Q(B^s)\theta_q(B)\varepsilon_t + \zeta X_t \qquad (9)$$

Where $B^s$ is the seasonal reverse potential operator; $\Phi_p$ is the seasonal autoregressive coefficient; $\Theta_Q$ is the seasonal moving average coefficient; $\zeta$ is the coefficient of exogenous variable. Other parameters are the same as ARIMA model ([32]; [50]). In this paper, SARIMAX(0,1,3) model is used for $CO_2$ emissions prediction.

### 3.3.4 Artificial Neural Network (ANN)

ANN refers to the complex network structure formed by the interconnection of a large number of processing units (neurons), which is a kind of abstraction, simplification and simulation of the organizational structure and operating mechanism of the human brain. ANN simulates neuronal activity with mathematical models and it is an information processing system based on the imitation of the structure and function of the brain neural network. ANN model has self-learning, self-organization, self-adaptation, strong nonlinear function approximation ability and strong fault tolerance ability, therefore it becomes a powerful tool to deal with various data types (nonlinear, fuzzy, noise) in machine learning algorithms [34]. For ANN, there is no formula that says what structure or which learning algorithm to use to find the optimal solution. Therefore, the best solution is obtained by trial and error [49].

ANN is usually divided into single layer and multi-layer. However, multi-layer ANN is often used in artificial neural networks since it has stronger imitation ability of biological neurons [57]. The ANN structure, as shown in Figure 3, usually consists of three parts: input layer, hidden layer and output layer.

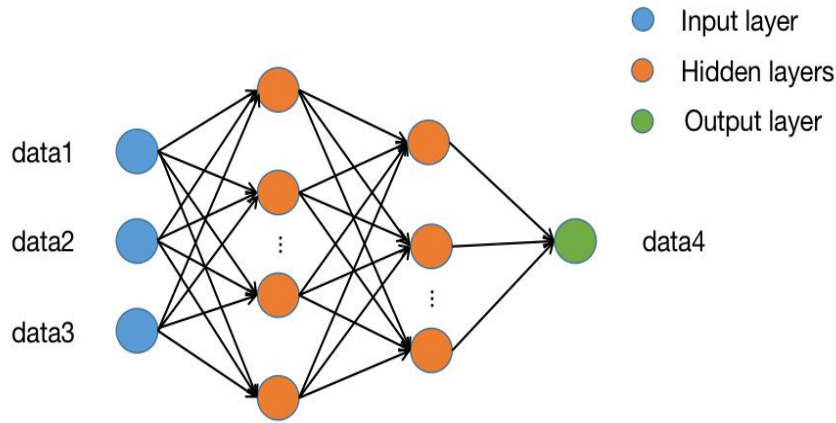

Fig.3. Architectural structure of ANN model.

In this paper, the $CO_2$ emissions data of the first three days will be used as input values, and the data of the fourth day will be used as output values to train the ANN model. See Table 2 for parameters setting in ANN model.

Table 2 parameters for ANN model.

| Parameters | Details |
| --- | --- |
| Input layer | Neure: 3 |
| Hidden layer 1 | Dense layer<br>Neure: 12<br>Activation: relu |
| Hidden layer 2 | Dense layer<br>Neure: 12<br>Activation: relu |
| Output layer | Dense layer<br>Neure: 1<br>Activation: linear |
| Loss function | Mean squared error loss |
| Optimizer | Adam |
| Epochs | 3000 |
| Batch size | 32 |
| Total parameters | 217 |

### 3.3.5 Random Forest (RF)

RF is an integrated learning algorithm based on decision tree, which was first proposed by [33]. As RF model shows amazing performance in classification and regression, it is often hailed as "the approach that represents the technological level of integrated learning" [56]. The modeling process is as follows:

Step 1: m samples are randomly taken out from the original data to generate m training sets.

Step 2: m decision tree models are trained from m training sets respectively.

Step 3: For a single decision tree model, it is assumed that the number of training sample features is n, and the best features are selected for splitting according to information gain/information gain ratio/Gini coefficient.

Step 4: The generated decision trees are formed into a random forest. For the classification problem, the final classification result is decided by the vote of multiple tree classifiers. For the regression problem, the mean value of multiple trees determines the final prediction result.

The parameters for RF model used in this paper can be seen in Table 3.

Table 3 parameters for RF model.

| Parameters | Values |
| --- | --- |
| N_estimators | 100 |
| Max_depth | 20 |
| Random_state | 2 |
| Criterion | MSE |

### 3.3.6 Long Short Term Memory (LSTM)

LSTM is a special recurrent neural network (RNN). With the increase of training time and the number of network layers, the traditional RNN is prone to gradient explosion or disappearance in the training process. To solve this problem, the LSTM model adds three logical structures to the RNN: input gate, forget gate, and output gate. The structure diagram of the model is shown in Figure 4. In LSTM model, the input gate, forget gate and output gate are denoted by i, f and o respectively; the memory unit is represented by C; the input data is X; the state of the unit is H. Input gate is used to determine how much of the network's input data needs to be saved to the cell state at the current time. Forget gate is used to remove all the less important information about the cell's state. The output gate is used to select the output of valuable cell states.

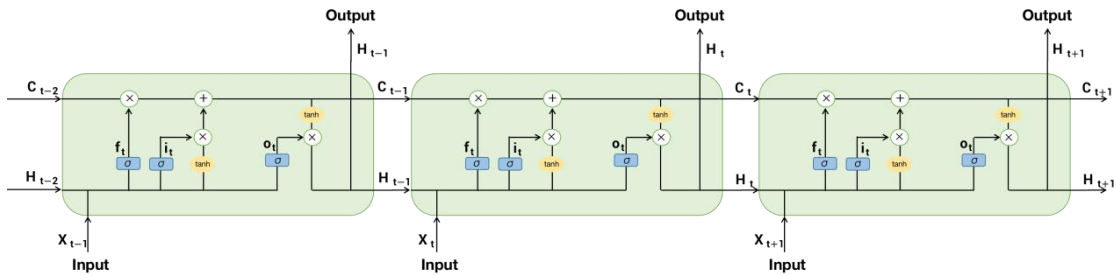

Fig.4. Architectural structure of LSTM model.

The following describes the modeling process of the LSTM model.

Step 1: Identify information that neurons need to forget. The formula for calculating the forget gate at time t is as following equation (10):

$$f_t = \sigma(W_f[H_{t-1}, X_t] + b_f) \tag{10}$$

where $\sigma$ is the sigmoid function; $W_f$ is the parameter to be determined; $b_f$ is a bias.

Step 2: Identify information that needs to be updated and put it into the cell state. This process consists of two steps: (1) Determine what information needs to be updated by the sigmoid function. (2) Generate a new candidate value through the tanh function. The expression is as equations (11) and (12):

$$i_t = \sigma(W_i[H_{t-1}, X_t] + b_i) \tag{11}$$

$$\bar{C}_t = \tanh(W_c[H_{t-1}, X_t] + b_c) \tag{12}$$

where the tanh function is the hyperbolic tangent function; $W_i$ and $W_c$ are the weights to be determined; $b_i$ and $b_c$ are biases.

Step 3: Update memory status. According to the part $f_t$ saved by the forget gate dot $C_{t-1}$, and add the updated parts $i_t$ dot $\bar{C}_t$. Finally get the updated memory status $C_t$. The calculation formula is as equation (13):

$$C_t = f_t * C_{t-1} + i_t * \bar{C}_t \tag{13}$$

According to the above equation, as the forget gate $f_t$ approaches 1 and input gate $i_t$ approaches 0, the storage unit in the old state is saved to the current time. LSTM uses this method to solve the problem of gradient disappearance in the RNN [53].

Step 4: Determine the output information. The calculation formula of the output gate at time t is equation (15):

$$O_t = \sigma(W_o[H_{t-1}, X_t] + b_o) \tag{14}$$

$$H_t = O_t * \tanh(C_t) \tag{15}$$

where $W_o$ is the weight to be determined and $b_o$ is the bias.

The parameters for LSTM model used in this paper can be seen in Table 4.

Table 4 parameters for LSTM model.

| Parameters | Details |
| --- | --- |
| The number of layers | 3 |
| First layer | LSTM layer<br>Nodes: 50<br>Activation: relu |
| Second layer | LSTM layer<br>Nodes: 50<br>Activation: relu |
| Third layer | Dense layer<br>Nodes: 1<br>Activation: linear |
| Loss function | Mean squared error loss |
| Optimizer | Adam |
| Epochs | 3000 |
| Batch size | 32 |

### 3.4 Evaluation criteria

In this section, we will discuss the evaluation criteria to explore the performance of the proposed models used to predict $CO_2$ emissions, which is a kind of regression problem. Here we collect five evaluation criteria to evaluate the models' performance which are: Mean Squared Error (MSE), Root Mean Squared Error (RMSE), Mean Absolute Error (MAE), Mean Absolute Percentage Error (MAPE), Coefficient of Determination ($R^2$). These statistical indicators are defined as follows:

**(1) Mean Squared Error (MSE)**

$$\text{MSE} = \frac{1}{n}\sum_{i=1}^{n}(y_i - \hat{y}_i)^2 \tag{16}$$

**(2) Root Mean Squared Error (RMSE)**

$$\text{RMSE} = \sqrt{\text{MSE}} \tag{17}$$

**(3) Mean Absolute Error (MAE)**

$$\text{MAE} = \frac{\sum_{i=1}^{n}|\hat{y}_i - y_i|}{n} \tag{18}$$

**(4) Mean Absolute Percentage Error (MAPE)**

$$\text{MAPE} = \frac{1}{n}\sum_{i=1}^{n}\left|\frac{\hat{y}_i - y_i}{y_i}\right| \times 100\% \tag{19}$$

**(5) Coefficient of Determination ($R^2$)**

$$R^2 = 1 - \frac{\sum_{i=1}^{n}(y_i - \hat{y}_i)^2}{\sum_{i=1}^{n}(y_i - \frac{\sum_{j=1}^{n}y_j}{n})^2} \tag{20}$$

where $y_i$ and $\hat{y}_i$ represent the actual and prediction values of $CO_2$ emissions respectively and n denotes the number of test data.

## 4. Results and discussions

All models proposed in this paper are written in the python 3.10 programming environment and we also use Keras and TensorFlow to implement machine learning models. Table 5 shows the values calculated from the evaluation criteria to evaluate the performance of the proposed models in this research.

MSE condition states that the closer the MSE is to zero, the better the prediction is. The MSE values range from 3.5179e-04 to 0.0229. The results show that the three machine learning models work better than that statistical models on MSE values. The LSTM model has the lowest MSE value of 3.5179e-04, followed by ANN model with the value of 3.9956e-04, and RF with the value of 5.3638e-04.

RMSE is the square root of MSE, so the model with positive values in all cases. The closer RMSE value to zero, the more accuracy the prediction results. The values of RMSE range from 0.0187 to 0.1516 and the LSTM model performs the best among other models on RMSE value since it has the 0.0187 value, which is the closest to zero. Followed by ANN with 0.0199 and RF with 0.0231.

MAE /MAPE indicates that the closer the value is to zero, the better the model is. Based on the value of MAE /MAPE of all models, it can be seen that LSTM model has presented the best performance than other models with its value 0.0140/14.8291%, which is the lowest and closest to zero. Followed by ANN and RF with values 0.0143/15.7681% and 0.0185/16.2964% respectively.

The model with $R^2$ value closer to 1 is better in predicting. $R^2$ values range from -0.0430 to 0.9844 in the six models and all of the statistical models are negative (-0.0430, -0.0148 and -0.0082), which indicates that the predictions tend to be less accurate than the average values of the data. LSTM model outperformed others in $R^2$ value with the value 0.9844 which is the closest to 1.

Overall, all of the machine learning models perform better than the statistical models on the five evaluation criteria. The best model that fits the long daily time series data is LSTM, which performs better in terms of MSE, RMSE, MAE, MAPE, and $R^2$, and followed by ANN and RF model. GM(1,1) model is the worst in terms of MSE, RMSE, and $R^2$, which makes it the least desirable model for long daily time series data predicting.

Table 5 Calculated values of evaluation criteria.

| Evaluation Criteria | Statistical Models | | | Machine Learning Models | | |
|---|---|---|---|---|---|---|
| | GM(1,1) | ARIMA | SARIMAX | RF | ANN | LSTM |
| MSE | 0.0229 | 0.0223 | 0.0222 | 5.3638e-04 | 3.9956e-04 | **3.5179e-04** |
| RMSE | 0.1516 | 0.1495 | 0.1490 | 0.0231 | 0.0199 | **0.0187** |
| MAE | 0.1318 | 0.1223 | 0.1247 | 0.0185 | 0.0143 | **0.0140** |
| MAPE(%) | 120.4531 | 138.9119 | 134.9591 | 16.2964 | 15.7681 | **14.8291** |
| $R^2$ | -0.0430 | -0.0148 | -0.0082 | 0.9763 | 0.9823 | **0.9844** |

In the same time, we have plotted the graph for the actual and prediction $CO_2$ emissions on the test data as shown in Figure 5 where the blue curve is actual value and the orange curve is the prediction value by model.

From Fig.5. (a), the prediction values of GM(1,1) model is characterized by a slow linear growth trend which has a huge difference compared with test data. From (b), we can see that

ARIMA's predicting values have a fluctuation in the initial position, after which it is almost a horizontal line. SARIMAX predicting result is almost the same pattern as ARIMA can be seen from Fig.5. (c), where the difference is that it has a certain fluctuation. Obviously, the difference between the predicting values and the actual values of GM(1,1)/ARIMA/SARIMAX model are too large and they can not reflect the trend on the test set well. Since GM(1,1)/ARIMA/SARIMAX model needs work on stationary data, so differential processing of the data is required, which is precisely the reason for the loss information and inaccurate result. From Table 5, we can see that the GM(1,1)/ARIMA/SARIMAX model has a MAPE value of 120.4531%/138.9119%/134.9591%, which is inaccurate for predicting. With the $R^2$ values of -0.0430/-0.0148/-0.0082, we can conclude that predicting with these three models will be inappropriate for the long daily time series data.

Based on ML models (ANN, RF and LSTM), the fitting curves of these models in $CO_2$ emissions prediction on test data are shown in Fig.5 (d)-(f) and it can apparently draw the conclusion that the fitting curve of ML models are superior to statistical models. In the same time, Fig.5 certifies that ML models' simulation for the total volume of $CO_2$ emissions in China offers excellent results. The prediction of ML models follow the historical values with very high accuracy which powerfully supports the idea that the $CO_2$ emissions can be successfully predicted by ML models. Furthermore, when we see the evaluation criteria values, we can see that the LSTM is the best performing model for predicting the near-real-time daily $CO_2$ emissions and it can give quite good and satisfactory results.

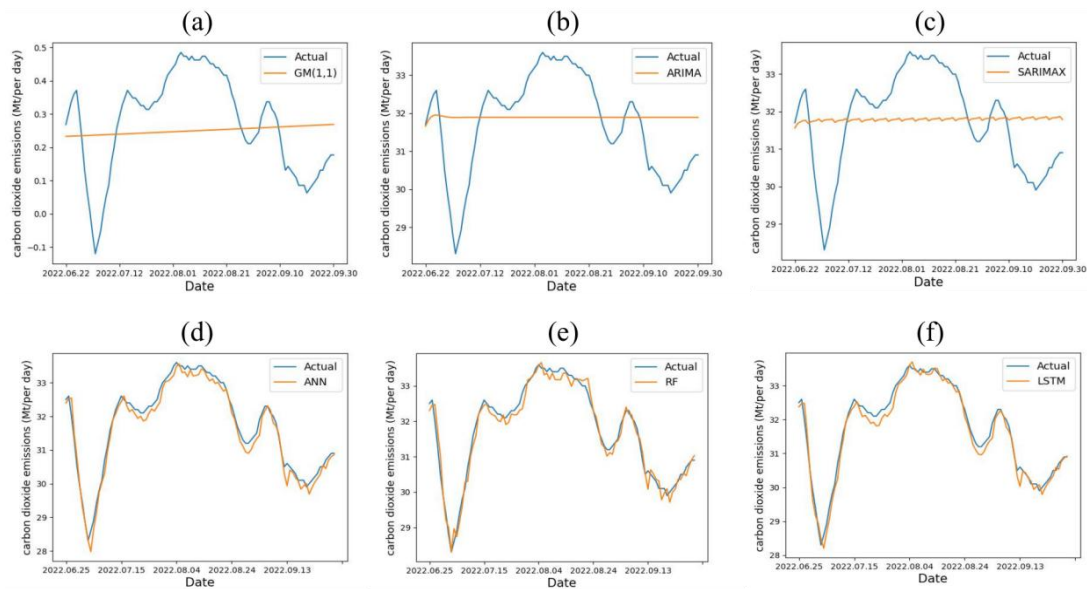

Fig.5. The fitting curves of six models in $CO_2$ emissions prediction on test sets.

As shown in Fig.6, the predicting results based on LSTM model of $CO_2$ emissions on test data (orange curve) from June 25st, 2022 to September 30st, 2022 and in the next three months (green curve) from October 1st, 2022 to December 31st, 2022 are presented. According to the predicting results, the values of carbon emissions for the next three months follow a similar trend to historical data.

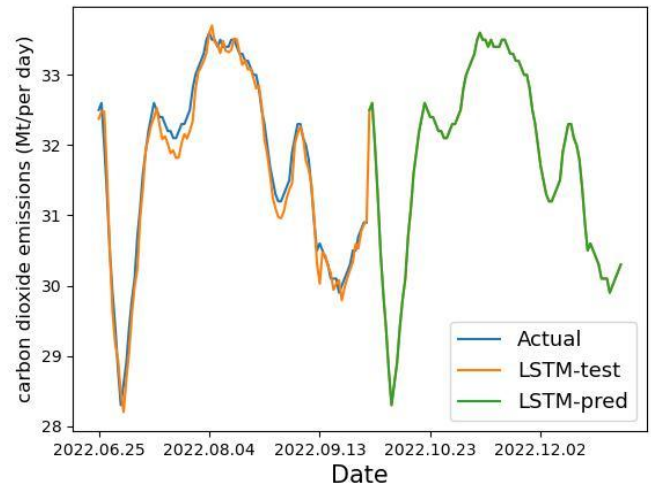

Fig.6. Testing and predicting of $CO_2$ emissions with respect to LSTM model.

## 5. Conclusions and policy implications

The near-real-time daily accurate prediction of carbon dioxide emissions in China is crucial for the government to control the $CO_2$ emissions for mitigating climate change. This paper discusses the performance of three statistical models (GM(1,1), ARIMA and SARIMAX) and three ML models (ANN, RF and LSTM) for near-real-time daily prediction of $CO_2$ emissions in China based on the near-real-time series data between January 1st, 2020 and September 30st, 2022 of all sectors (Power、Industry、Ground Transport、Residential、Domestic Aviation、International Aviation) in China. In order to evaluate the performance of these models and choose the best suitable one for future predicting, five criteria (MSE, RMSE, MAE, MAPE and $R^2$) are discussed in detail.

Results indicate that all of the proposed models in this paper, ML models perform better than statistical models for daily time series $CO_2$ emissions prediction. To analyze the performance of the proposed models, we have used 1004 daily $CO_2$ emissions data for modeling and the performance evaluation indicates that LSTM is the best performing model on all five criteria with a 3.5179e-04 MSE value, 0.0187 RMSE value, 0.0140 MAE value, 14.8291% MAPE value, and 0.9844 $R^2$ value. Therefore, the LSTM model is used to predict the next 3 months' $CO_2$ emissions trend.

In the future, as Carbon Monitor project continues to update the daily carbon emissions data of China by projecting emissions in real time, we can detect subtle changes and make timely policy adjustments accordingly. For example, in the post-COVID-19 pandemic, various activities will be resumed in all regions to restore economic development. But if economic recovery and stimulus are not monitored, $CO_2$ emissions would quickly rebound and rise above pre-pandemic levels, which would be adverse to our target of carbon neutrality. It is for this reason that near-real-time $CO_2$ emissions predicting is so important to policymaking. At the same time, timely predicting of carbon dioxide emissions will facilitate timely policy adjustment and management during the ongoing energy transition. Thus it is important to detect changes in $CO_2$ emissions in a timely manner in order to reduce the time required for policy adjustment (perhaps 1-2 years shorter than annual emissions predicting). As a result, it is strongly recommended that governments implement policies and legislations to reduce $CO_2$ emissions in all sectors.

・Government needs to advocate low-carbon living and raise the awareness of the whole people about energy conservation.

・Green development of ground transport infrastructure and steady promotion of the popularity of electric vehicles will help reduce carbon emissions in the transport sector.

・The aviation sector can reduce emissions by using green fuels and building green airports.

・China should vigorously develop renewable energy sources, like solar power, geothermal, tidal waves, wind power, and biomass for its power supply and reduce its dependence on coal.

The flaw in this work is only to univariate time series data prediction for $CO_2$ emissions. We have not inculcated many influencing factors such as GDP, population, energy consumption, future policy and so on, which are the exogenous factors that can be explored in the future. By considering the above factors, we can find the correlation between them and carbon dioxide emissions, and conduct multivariate analysis, so as to give more effective $CO_2$ emissions prediction methods and policy recommendations.

## Acknowledgements

We appreciate to the persons who gave valuable comments that helped bring considering improvements to the paper significantly. This work is supported by Capital University of Economics and Business (no. XRZ2022064).

## References

[1]L. Abdullah, H.M. Pauzi, Methods in forecasting carbon dioxide emissions: a decade review, Jurnal. Teknologi., 75 (2015) 67–82.

[2]M. Abunofal, N. Poshiya, R. Qussous, A. Weidlich, Comparative Analysis of Electricity Market Prices Based on Different Forecasting Methods, 14th IEEE Madrid PowerTech Conference (IEEE POWERTECH), (2021).

[3]Akay, M. Atak, Grey prediction with rolling mechanism for electricity demand forecasting of Turkey, Energy, 32 (2007) 1670e5.

[4]Alam, A. AlArjani, A Comparative Study of CO2 Emission Forecasting in the Gulf Countries

Using Autoregressive Integrated Moving Average, Artificial Neural Network, and Holt-Winters Exponential Smoothing Models, ADV. METEOROL., (2021) 8322590.

[5]Amarpuri, N. Yadav, G. Kumar, S. Agrawal, Prediction of CO2 emissions using deep learning hybrid approach: a case study in indian context, In: 2019 twelfth international conference on contemporary computing (IC3) IEEE, (2019) 1–6.

[6]S.A. Bandh, S. Shaf, M. Peerzada et al., Multidimensional analysis of global climate change: a review, Environ. Sci. Pollut. Res., 28 (2021) 24872–24888.

[7]M.A. Behrang, E. Assareh, M. Assari, R.A. Ghanbarzadeh, Using bees algorithm and artifcial neural network to forecast world carbon dioxide emission, Energy Sources, Part A: Recovery, Util Environ. Efects., 33 (2011), 1747–1759.

[8]O.F. Cansiz, K. Unsalan, F. Unes, Prediction of CO2 emission in transportation sector by computational intelligence techniques, Int. J. Global Warm., 27 (2022) 271-283.

[9]S. Chaabouni, Modeling and forecasting 3E in Eastern Asia: a comparison of linear and nonlinear models, Qual. Quant., 50 (2016) 1993-2008.

[10]Chiroma, S. Abdul-kareem, A. Khan et al., Global warming: predicting OPEC carbon dioxide emissions from petroleum consumption using neural network and hybrid cuckoo search algorithm, PLoS. One., 10 (2015) e0136140.

[11]Y. Cho, J.C. Hwang, C.S. Chen, Customer short term load forecasting by using ARIMA transfer function model, In: IEEE conference on energy management and power delivery, 1 (1995) 317–322.

[12]B. Cortez, B. Carrera, Y.J. Kim, J.Y. Jung, An architecture for emergency event prediction using LSTM recurrent neural networks, Expert. Syst. Appl., 97 (2018) 315–324.

[13]J. Crespo, Cuaresma, J. Hlouskova, S. Kossmeier, M. Obersteiner, Forecasting electricity spot-prices using linear univariate timeseries models, Appl. Energy, 77 (2004) 87–106.

[14]J.L. Deng, Introduction to grey system theory, J. Grey Syst., 1 (1989) 1e24.

[15]J.L. Deng, The Basis of Grey Theory, Press of Huazhong University of Science & Technology, Wuhan, China, 2002.

[16]S. Ding, Y.G. Dang, X.M. Li, J.J. Wang, K. Zhao, Forecasting Chinese CO2 emissions from fuel combustion using a novel grey multivariable model, J. Clean Prod., 162 (2017) 1527-1538.

[17]K. Dong, X. Dong, C. Dong, Determinants of the global and regional CO2 emissions: what causes what and where? Appl. Econ., 51 (2019) 5031–5044.

[18]H.M. Duan, W.G. Nie, A noval grey model based on Susceptible Infected Recovered Model: A case study of COVD-19, Physica A: Statistical Mechanics and its Applications, 602 (2022) 127622.

[19]X. Fang, W. Liu, J. Ai, M. He, Y. Wu, Y. Shi, W. Shen, C. Bao, Forecasting incidence of infectious diarrhea using random forest in Jiangsu Province, China BMC Infectious Diseases, 20 (2020), 1–8.

[20]P. Gopu, R.R. Panda, N.K. Nagwani, Time series analysis using ARIMA model for air pollution prediction in Hyderabad city of India, In: In Soft Computing and Signal Processing, Springer Singapore, (2021) 47–56.

[21]C. Hamzacebi, I. Karakurt, Forecasting the Energy-related CO2 Emissions of Turkey Using a

Grey Prediction Model, Energy Sources Part A-Recovery Util. Environ. Eff., 37 (2015) 1023-1031.

[22] Y. Hao, H. Chen, Y.M. Wei, Y.M. Li, The infuence of climate change on CO2 (carbon dioxide) emissions: an empirical estimation based on Chinese provincial panel data, J. Clean. Prod., 131 (2016) 667–677.

[23] L.C. Hsu, Technological Forecasting and Social Change, 70 (2003) 563-574.

[24] Y. Huang, L. Shen, H. Liu, Grey relational analysis, principal component analysis and forecasting of carbon emissions based on long short-term memory in China, J. Clean. Prod., 209 (2019) 415–423.

[25] IEA, Energy Technology Perspectives 2020, Special Report on Carbon Capture Utilization and Storage, Paris, International Energy Agency, (2020).

[26] IPCC, Climate change 2014: synthesis report. Contribution of working groups I, II, III to the fifth assessment report of the intergovernmental panel on climate change, IPCC, Geneva, 151 (2014).

[27] H.C. Jeong, J. Jung, B.O. Kang, Development of ARIMA- based forecasting algorithms using meteorological indices for seasonal peak load, Trans Korean Inst. Electr. Eng., 67 (2018) 1257–1264.

[28] S.A. Kalogirou, Applications of artifcial neural-networks for energy systems, Appl. Energy, 67 (2000) 17–35.

[29] B. Kordanuli, L. Barjaktarović, L. Jeremić, M. Alizamir, Appraisal of artificial neural network for forecasting of economic parameters, Physica A: Statistical Mechanics and its Applications, 465 (2017) 515-519.

[30] M. Kour, Modelling and forecasting of carbon-dioxide emissions in South Africa by using ARIMA model, INT. J. ENVIRON. SCI. TE., DOI: 10.1007/s13762-022-04609-7 (2022).

[31] S. Kumari, S.K. Singh, Machine learning-based time series models for efective CO2 emission prediction in India, Environ. Sci. Pollut. Res., (2022).

[32] G.C. Lee, J. Han, Forecasting gas demand for power genera- tion with SARIMAX models, Korean Manag. Sci. Rev., (2020) 67–78.

[33] B. Leo, Random Forest, Machine Learning, 45 (2001) 5-32.

[34] R. Lippmann, An introduction to computing with neural nets, IEEE Assp magazine, 4 (1987) 4-22.

[35] Z. Liu, et al., Near-real-time monitoring of global CO2 emissions reveals the efects of the COVID-19 pandemic, Nat. Commun, 11 (2020a) 5172 .

[36] Z. Liu et al., Carbon Monitor, a near-real-time daily dataset of global CO2 emission from fossil fuel and cement production, Sci. Data, 7 (2020b) 392 .

[37] L.W. Liu, H.J. Zong, E.D. Zhao, C.X. Chen, J.Z. Wang, Can China realize its carbon emission reduction goal in 2020: From the perspective of thermal power development, Appl. Energy, 124 (2014) 199- 212.

[38] L. Milačić, S. Jović, T. Vujović, J. Miljković, Application of artificial neural network with extreme learning machine for economic growth estimation, Physica A: Statistical Mechanics and its Applications, 465 (2016) 285-288.

[39] M.K.A. Nishan, V.M. Ashiq, Role of energy use in the prediction of CO2 emissions and economic growth in India: evidence from artificial neural networks (ANN), Environ. Sci. Pollut. Res.,

27 (2020) 23631-23642.

[40] C. Nontapa, C. Kesamoon, N. Kaewhawong, P. Intrapaiboon, A new time series forecasting using decomposition method with SARIMAX model, In: International Conference on Neural Information Processing, Springer, Cham., (2020) 743–751.

[41] K. Nursac, O. Oktay, B. Murat, Estimation of Gas Emission Values on Highways in Turkey with Machine Learning, 10th IEEE International Conference on Renewable Energy Research and Applications, (2021) 443-446.

[42] T. Nyoni, W.G. Bonga, Prediction of CO2 emissions in india using ARIMA models, DRJ-J Econ Finance, 4 (2019) 1–10.

[43] H.T. Pao, C.M. Tsai, Modeling and forecasting the CO2 emissions, energy consumption, and economic growth in Brazil, Energy, 36 (2011) 2450 - 2458.

[44] A. Rehman, H. Ma, M. Ahmad, et al., Towards environmental Sustainability: Devolving the infuence of carbon dioxide emission to population growth, climate change, Forestry, livestock and crops production in Pakistan, Ecol. Indic., 125 (2021) 107460.

[45] J.K. Sethi, M. Mittal, Analysis of Air Quality using Univariate and Multivariate Time Series Models, 10th International Conference on Cloud Computing, Data Science and Engineering (Confluence), (2020) 823-827.

[46] A. Sozen, Z. Gülseven, E. Arcaklioglu, Forecasting based on sectoral energy consumption of GHGs in Turkey and mitigation policies, Energy Pol., 35 (2007) 6491e6505.

[47] M. Stevanović, S. Vujičić, A.M. Gajić, Gross domestic product estimation based on electricity utilization by artificial neural network, Physica A: Statistical Mechanics and its Applications, 489 (2018) 28-31.

[48] W. Sun, Y.W. Wang, C.C. Zhang, Forecasting CO2 emissions in Hebei, China, through moth-flame optimization based on the random forest and extreme learning machine, Environ. Sci. Pol., 25 (2018) 28985–28997.

[49] A.E. Tumer, A. Akkus, Forecasting Gross Domestic Product per Capita Using Artificial Neural Networks with Non-Economical Parameters, Physica A: Statistical Mechanics and its Applications, 512 (2018) 468-473.

[50] S.I. Vagropoulos, G.I. Chouliaras, E.G. Kardakos, C.K. Simoglou, A.G. Bakirtzis, 'Comparison of SARIMAX, SARIMA, modified SARIMA and ANN-based models for short-term PV generation forecasting', in 2016 IEEE International Energy Conference (ENERGYCON) April, (2016).

[51] Z.X. Wang, D.D. Li, H.H. Zheng, Model comparison of GM(1,1) and DGM(1,1) based on Monte-Carlo simulation, Physica A: Statistical Mechanics and its Applications, 542 (2020) 123341.

[52] Z.X. Wang, D.J. Ye, Forecasting Chinese carbon emissions from fossil energy consumption using non-linear grey multivariable models, J. Clean. Prod., 142 (2017) 600–612.

[53] X. Xu, Y. Ding, S.X. Hu, et al., "Scaling for edge inference of deep neural networks," Nature Electronics, 1 (2018) 216–222.

[54] A.W.L. Yao, S.C. Chi, C.K. Chen, Development of an integrated grey-fuzzy-based electricity management system for enterprises, Energy, 30 (2005) 2759e71.

[55] A.W.L. Yao, S.C. Chi, Analysis and design of a taguchi-grey based electricity demand

predictor for energy management systems, Energy Convers Manage, 45 (2004) 1205e17.

[56]Y. Ying, H. Wang, Dynamic random regression forests for real-time head pose estimation, Machine Vision and Applications, 24 (2013) 1705-1719.

[57]Zeinalizadeh, A.A. Shojaie, M. Shariatmadari, Modeling and analysis of bank customer satisfaction using neural networks approach, International Journal of Bank Marketing, 33 (2015) 717-732.

[58]X.W. Zhao, P. Zhao, L.W. Zhu, G.Y. Zhang, A Comparison of Multivariate and Univariate Time Series Models Applied in Tree Sap Flux Analyses, Forest science, 68 (2022) 473-486.

[59]B.W. Zhou, K.L. Ang, A. Poh, trigonometric grey prediction approach to forecasting electricity demand, Energy, 31 (2006) 2839e47.

[60]W.H. Zhou, B. Zeng, X.Z. Liu, Forecasting Chinese carbon emissions using a novel grey rolling prediction model, Chaos Solitons Fract, 147 (2021).

[61]Z. Zuo, H. Guo, J. Cheng, An LSTM-STRIPAT model analysis of China's 2030 $CO_2$ emissions peak, Carbon Manag, 11 (2020) 577–592.